\documentclass[10pt,twocolumn,letterpaper]{article}

\usepackage[pagenumbers,applications]{wacv}      

\usepackage{graphicx}
\usepackage{amsmath}
\usepackage{amssymb}
\usepackage{booktabs}
\usepackage{multirow}
\usepackage[pagebackref,breaklinks,colorlinks]{hyperref}
\usepackage{graphicx}
\usepackage{hyperref}
\usepackage{xcolor}

\usepackage[capitalize]{cleveref}
\crefname{section}{Sec.}{Secs.}
\Crefname{section}{Section}{Sections}
\Crefname{table}{Table}{Tables}
\crefname{table}{Tab.}{Tabs.}

\def\wacvPaperID{2} 
\def\confName{WACV}
\def\confYear{2025}

\newcommand{\ours}{Mediffusion}

\begin{document}

\title{Mediffusion: Joint Diffusion for Self-Explainable Semi-Supervised Classification and Medical Image Generation}

\author{Joanna Kaleta\\
Warsaw University of Technology\\
Sano Centre for Computational Medicine\\
{\tt\small joanna.kaleta.dokt@pw.edu.pl}
\and
Paweł Skierś\\
Warsaw University of Technology\\
{\tt\small pawel.skiers.stud@pw.edu.pl}
\and
Jan Dubiński\\
Warsaw University of Technology\\
IDEAS NCBR\\
{\tt\small jan.dubinski.dokt@pw.edu.pl}
\and
Przemysław Korzeniowski\\
Sano Centre for Computational Medicine\\
{\tt\small p.korzeniowski@sanoscience.org}
\and
Kamil Deja\\
Warsaw University of Technology\\
IDEAS NCBR\\
{\tt\small kamil.deja@pw.edu.pl}
}

\maketitle

\begin{abstract}
We introduce \ours{} -- a new method for semi-supervised learning with explainable classification based on a joint diffusion model. The medical imaging domain faces unique challenges 
due to scarce data labelling -- insufficient for standard training, and critical nature of the applications that require high performance, confidence, and explainability of the models. In this work, we propose to tackle those challenges with a single model that combines standard classification with a diffusion-based generative task in a single shared parametrisation. By sharing representations, our model effectively learns from both labeled and unlabeled data while at the same time providing accurate explanations through counterfactual examples. In our experiments, we show that our \ours{} achieves results comparable to recent semi-supervised methods while providing more reliable and precise explanations.
\end{abstract}

\section{Introduction}
\label{sec:intro}

Deep learning methods achieve remarkable performance across various computer vision domains, including medical image analysis \cite{Celard2023}. However, this field faces unique challenges compared to other vision tasks due to the scarcity of labeled data and the need for methods explainability \cite{challenges, manifesto}. Annotating medical images requires specialized expert knowledge, making the labeling process costly and time-consuming. This often results in datasets that are insufficient in size for effective machine learning training. Moreover, the nature of healthcare imaging applications often demands that the final solutions be transparent and interpretable, as decisions made on their basis can have severe consequences \cite{manifesto}. 

In this work, we propose to tackle those two problems with Denoising Diffusion Probabilistic Models (DDPM)~\cite{sohl2015deep,ho2020denoising}. Applications based on those methods have already established new state-of-the-art in many domains such as image~\cite{dhariwal2021diffusion}, video~\cite{ho2022imagen}, music~\cite{liu2021diffsinger} or speech~\cite{popov2021grad} generation. In the field of medical imaging, the same methods are used for synthetic image generation~\cite{kim2022diffusion} or their reconstruction~\cite{xie2022measurement, kaleta2024}.

\begin{figure}[t!]
    \centering
    \includegraphics[width=\columnwidth]{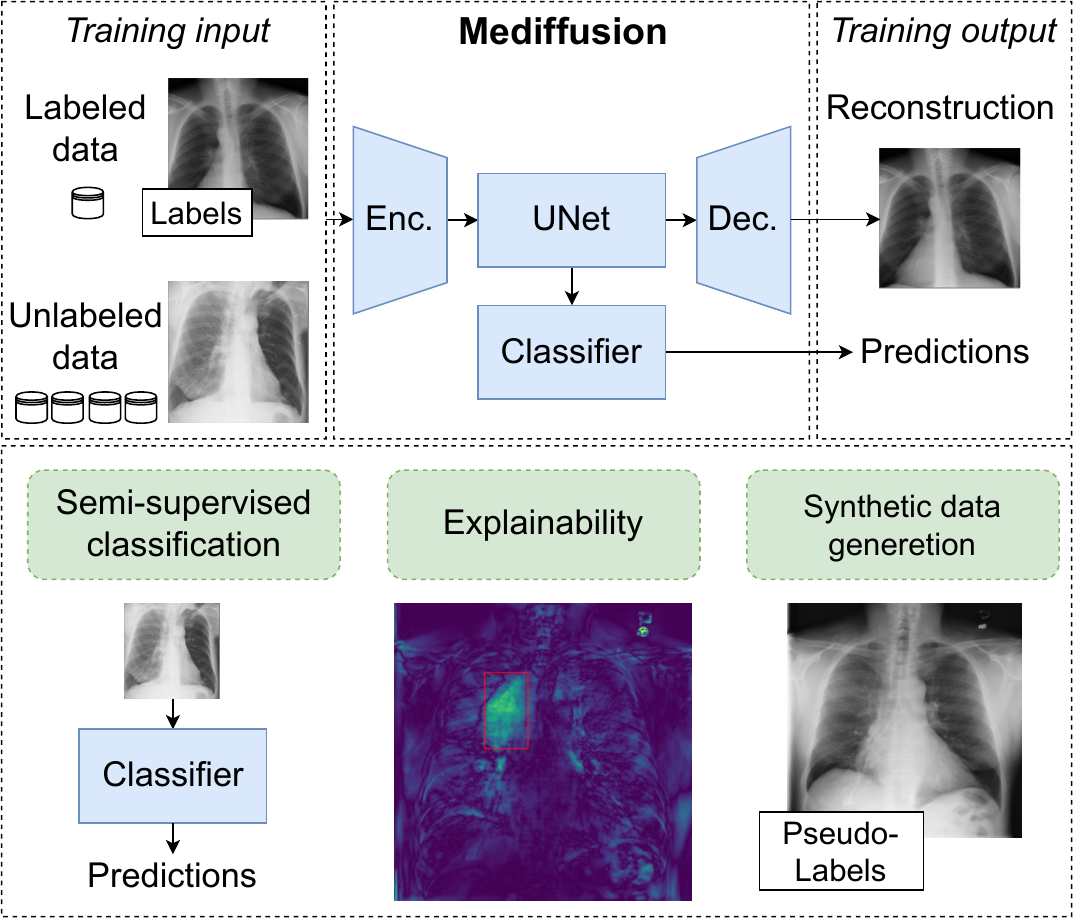}
    \caption{\textbf{\ours{}  training and capabilities.} Our proposed method utilizes both labeled and unlabeled data samples to build joint representations suitable for generative and discriminative tasks. We evaluate our method in 3 tasks (1) semi-supervised classification (2) inherent visual explainability of classifier decision (3) synthesising new pseudo-labeled data samples. }
    \label{fig:teaser}
    \vspace{-0.5cm}
\end{figure}

Nevertheless, apart from the astonishing performance in synthesizing new data samples, DDPMs can also be used as building blocks for other downstream tasks such as classification~\cite{yang2023diffmic}, image segmentation~\cite{baranchuk2021labelefficient} or correspondence~\cite{luo2024diffusion}, with some extensions to explainable methods~\cite{augustin2022diffusion}.

In this work, we propose to benefit from the representations learned by diffusion models in the medical domain. In particular, we introduce \ours{} -- a joint diffusion model that uses a shared parametrization in the form of UNet~\cite{unet_ronneberger2015unet} model to solve both generative and discriminative tasks. We show how generative task with a diffusion objective can improve the performance of the model trained in the semi-supervised regime. Moreover, thanks to the shared features between the discriminative and generative models, we can provide accurate explanations for the decisions of the classifier by generating counterfactual examples. Such explanations are crucial in the application of ML methods in medical domain, where the decision of whether to use a model or not is often based on the trust in the system.

In our experiments, we show that \ours{} achieves performance comparable to recent state-of-the-art semi-supervised techniques while at the same time bringing accurate explainability and the possibility for sampling synthetic examples. The contributions of this work can be summarized as follows:
\begin{itemize}
    \item We introduce \ours{} -- a new method for joint medical data modeling with shared parametrization between generative and discriminative tasks.
    \item We show how we can improve the performance of the medical classifier in a semi-supervised setup through the diffusion-based generative task.
    \item We present the method for providing explanations to the classifier decisions through counterfactual generations.
\end{itemize}



\section{Background}
\label{sec:background}

\subsection{Diffusion models in the medical domain}
Diffusion models have proven useful for synthetic data generation and enhancement in the medical field, including tasks such as denoising, super-resolution, and data reconstruction \cite{chen2024generalizable, zingarini2023m3dsynth, wang2023inversesr3dbrainmri}. Beyond data generation, diffusion models are also increasingly applied in segmentation, anomaly detection, and classification \cite{wolleb2022diffusion, yang2023diffmic}. Our work leverages diffusion models for semi-supervised learning, enhancing classification performance while supporting explainability. Please note the extended related work is included in ~\cref{app:full_related_work}.

\subsection{Semi-supervised learning for classification}
Semi-supervised learning (SSL) leverages both labeled and unlabeled data to improve model accuracy with limited annotations. Approaches like ACPL \cite{acpl_liu2022acpl} and S2MTS2 \cite{s2mts2_liu2021selfsupervised} employ techniques such as classifier ensembling and contrastive learning to enhance SSL performance. Our method, rather than relying solely on pseudo-labeling or consistency regularization, integrates a joint representation learning approach that benefits both generative and discriminative tasks.

\subsection{Visual Counterfactual Explanations}
Visual Counterfactual Explanations (VCE) provide interpretable insights into model decisions by generating images with minimal changes required to alter predictions. In the medical domain, VCE has facilitated clearer decision-making for clinicians \cite{explenetion1, explenetion2, explenetion3}. We extend this by utilizing diffusion models for counterfactual generation within a unified generative-discriminative framework, eliminating the need for external explanation models.

\section{Our method}

\subsection{Method overview} \ours{} leverages a joint training framework for a latent diffusion model and a classifier, effectively utilizing both labeled and unlabeled data.   

First, we train a self-supervised vision autoencoder on the entire dataset. Then, we train the U-Net based latent diffusion model with a classifier component. The encoder of U-Net serves as a feature extractor for the classifier component. U-Net and classifier are being optimized jointly in the following way. The diffusion model is optimized in a self-supervised manner, leveraging all available data points. Available labeled data is used to train the classifier through supervised learning. This enables \ours{} to construct rich data representations while the classifier efficiently maps these representations to final class predictions.

At the inference stage, our multitask model can both \textbf{generate} and \textbf{classify} data. During data generation, we can use the classifier to guide the denoising process toward a specific class. This enables two key tasks: 1) generating class-conditioned synthetic data, and 2) creating \textbf{visual counterfactual examples}. The latter helps explain classifier decisions. For instance, if the model predicts \textit{Cardiomegaly}, it can also modify the classified image showing how a \textit{non-Cardiomegaly} case would appear, highlighting the key differences between diseased and healthy tissue for the same patient. 
In the next sections, we explain the components of \ours{} in detail.


\subsection{Latent encoder and decoder}
Due to the high resolution of medical images, directly applying diffusion models, as demonstrated in \cite{jointdiffusion}, is impractical in this domain. Therefore, we employ the latent diffusion models~\cite{rombach2022highresolution}, where the diffusion process is introduced in the latent space of a vision autoencoder. More details are included in \cref{app:latencoder}.

\subsection{Joint Training of Diffusion Model and Classifier}

Our model builds on prior work in joint training of diffusion models and classifiers \cite{jointdiffusion, rombach2022highresolution}. In the Diffusion Denoising Probabilistic Model (DDPM) framework, we use a latent diffusion model where both the forward and reverse processes operate in a latent space. In the forward phase, Gaussian noise $\epsilon \sim \mathcal{N}(0, I)$ is incrementally added to the latent representation $z$ over diffusion steps $t \in [0, T]$ to produce noisy samples $z_t$. The reverse phase aims to reconstruct the original sample by gradually denoising these representations, parameterized by a UNet denoising model.

Our approach integrates a classifier with the diffusion model through a shared UNet architecture. The UNet’s encoder-decoder structure captures hierarchical features from the noisy latent representation at each diffusion step, denoted as \( \mathbf{H}_t \), which is used for both denoising and classification tasks. We form the classifier \( g_\omega \), which takes features \( \mathbf{H}_t \) and outputs a predicted class label \(\hat{y}\). This joint approach allows for simultaneous generative and discriminative modeling.

The combined model can be formulated as:
\[
p_{\nu,\psi,\omega}(z_{0:T}, y) = p_{\nu,\omega}(y|z_0) p_{\nu,\psi}(z_{0:T}),
\]
with its log likelihood given by:
\begin{equation}
\ln p_{\nu,\psi,\omega}(z_{0:T}, y) = \ln p_{\nu,\omega}(y|z_0) + \ln p_{\nu,\psi}(z_{0:T}).
\label{eq:joint_dist}
\end{equation}
The classification loss \( L_{\text{class}} \) uses cross-entropy and applies only to labeled data, while the diffusion loss \( L_{t,\text{diff}} \) minimizes the distance between the predicted and actual noise added to the latent sample at each step. Our final loss function combines these:
\[
L(\nu, \psi, \omega) = L_{\text{class}}(\nu, \omega) - \sum_{t=0}^T L_{t,\text{diff}}(\nu, \psi).
\]

\subsection{Classifier Guidance}

Classifier guidance \cite{dhariwal2021diffusion} directs diffusion sampling towards a specified class by leveraging gradients from a classifier. In our model, we naturally integrate classifier guidance through the joint classifier \( g_\omega \). During sampling, we adjust the noise prediction based on the class label \( y \), using the gradient to guide the generated sample towards this target.

Following \cite{dhariwal2021diffusion}, the noise prediction \(\hat{\epsilon}'(z_t)\) at diffusion step \( t \) is adjusted as:
\begin{equation}   
\hat{\epsilon}'(z_t) := \hat{\epsilon}_\theta(z_t) - \sqrt{1 - \bar{\alpha}_t} \nabla_{z_t} \log g_\omega(y|e_\nu(z_t)),
\end{equation}
where \(\hat{\epsilon}_\theta(z_t)\) is the original noise prediction from the denoising model, \(\bar{\alpha}_t\) is a scheduling parameter that controls the noise level at each step, and \(\nabla_{z_t} \log g_\omega(y|e_\nu(z_t))\) is the gradient of the log-probability of the target class \( y \) with respect to the latent representation \( z_t \).

This adjustment refines the sampling process, allowing the diffusion model to generate samples closer to the desired class \( y \).

\begin{figure}[t!]
    \centering
    \scalebox{0.85}{
    \includegraphics[width=\columnwidth]{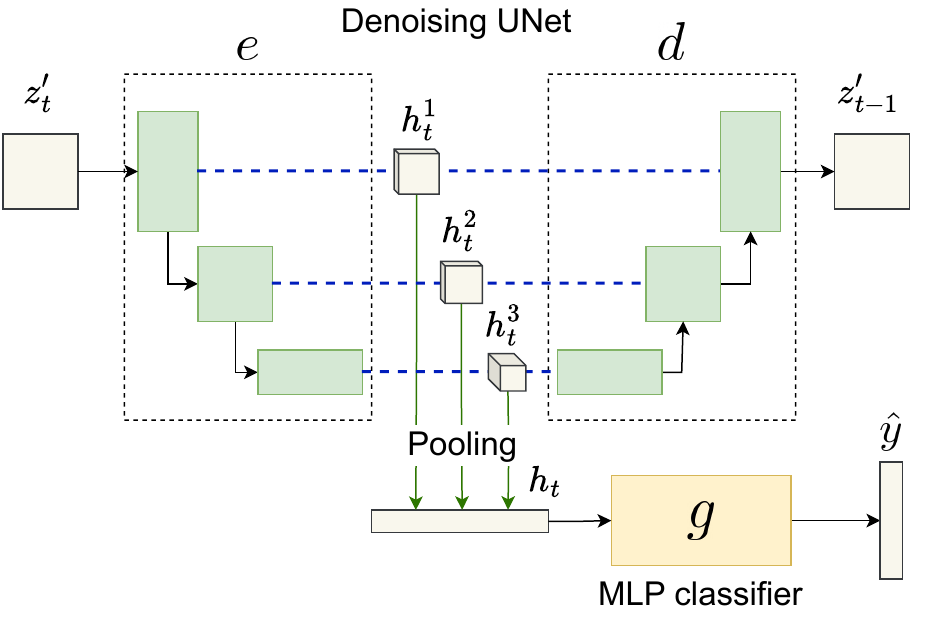}
    }
    \caption{\textbf{Joint training of diffusion model and classifier.} Data representation $h_t$ extracted in the UNet-based architecture of our joint latent diffusion model is utilized as input to the classifier component.}
    \label{fig:method}
    \vspace{-0.6cm}
\end{figure}

\subsection{Visual Counterfactual Explanations}
A unified model capable of generation and classification enables natural self-explanation of the classifier's decisions. With \ours, we can create counterfactual examples with the generative part of our
model by noising the original image and denoising it back
with guidance towards the opposite target. The objective is to explain the classifier’s decision by showing which modifications to the original images are needed to change the prediction. To obtain counterfactual examples for the classifier $g_\omega$, we use the following algorithm. First, we encode the image \( x \) into a latent representation \( z \) using the network \( \mathcal{E} \). Then, we obtain a noisy latent sample \( z_t \) where \( 0 < t < T \). In practice, to better preserve the original semantic content of the image $x$, we restrict \( t \) to \( 50 < t < 300 \). We then employ guided sampling towards \( z'_0 \) with the diffusion model, leveraging score-based classifier guidance either towards or against the desired prediction we aim to explain. 
By decoding the latent sample $z'_0$ using the decoder \( \mathcal{D} \), we arrive at an image in pixel space, denoted as \( x' = \mathcal{D}(z'_0) \). 


\section{Experiments and results}

In this section, we evaluate the performance of our \ours{} in three tasks. We first measure the performance of the classifier part of our model in both fully supervised and semi-supervised approaches. To that end, we conduct experiments on the  ChestXRay14 dataset \cite{nih_dataset} and ISIC2019 dataset \cite{isic_dataset}, where we compare our method with other state-of-the-art solutions. Next, we present a qualitative and quantitative evaluation of the counterfactual examples generation with \ours{}. Finally, we show the generative capabilities of our method.

\textbf{Datasets.} The  ChestXRay14 \cite{nih_dataset} dataset comprises 112,120 X-ray images from 30,805 unique patients, annotated with disease labels extracted using Natural Language Processing from radiological reports. The dataset includes 15 classes, covering 14 thoracic diseases and a "No findings" category, with bounding box annotations available for 8 of these classes. 
The ISIC2019 dataset \cite{isic_dataset} includes 25,331 dermoscopic images categorized into nine diagnostic classes and "None of the others." This dataset is designed for training and evaluating classification models in dermatology. ISIC2019 contains only images with corresponding labels and does not provide bounding boxes. 

\subsection{Semi-Supervised Classification}

Obtaining annotations and labels in the medical domain is particularly challenging due to the need for expertise from trained physicians. Consequently, unlabeled data is more prevalent than labeled data in this field, making semi-supervised learning highly relevant. We evaluate the performance of \ours{} and compare it against baselines and competing methods designed for this task. All competing methods utilize DenseNet-121 as their backbone. We conduct experiments with varying percentages of labeled data and report AUC values averaged across all classes.

\Cref{tab:nih_auc} presents the results for the classification task on the ChestXRay14 dataset. \ours{} achieves superior performance comparing to modern methods without pretraining, namely Densenet \cite{DenseNet2017}, NoTeacher \cite{noteacher_Unnikrishnan_2021} and ACPL \cite{acpl_liu2022acpl}. This highlights the capabilities of learning meaningful representations with our method in a semi-supervised manner, even without relying on a feature extractor pre-trained on a large external dataset. The benefits of \ours{} are most pronounced in setups with a smaller percentage of labeled data.

\begin{table}[h]
    \small
    \setlength{\tabcolsep}{4.4pt}
    \centering
    \caption{Performance comparison of different methods at various label percentages on ChestXRay14 dataset at official test split. We only show our reproduction without ImageNet pretrainig. Extended results including pretraining or values reported by other works are provided in \cref{app:results}.}
    \small
    \scalebox{1}{
    \begin{tabular}{llcccc}
        \toprule
        \multirow{2}{*}{Method Type} & \multirow{2}{*}{Method} & \multicolumn{4}{c}{Label Percentage} \\
        \cmidrule(r){3-6}
        & & 2\% & 5\% & 10\% &  20\% \\
        \midrule
        \multirow{1}{*}{Baseline} 
        & Densenet \cite{DenseNet2017}& 64.42 & 66.33  & 69.48 &  74.07   \\
        
        \midrule
        \multirow{1}{*}{Consistency} 
        &S2MTS2\cite{s2mts2_liu2021selfsupervised} & 62.22 &65.97&71.01&74.78\\
        \midrule
        \multirow{1}{*}{Pseudo Label}
        & ACPL ~\cite{acpl_liu2022acpl} & 63.22 & 66.68 & 71.79 &  74.93 \\
        \midrule
        \multirow{1}{*}{Diffusion} & \textbf{\ours{}} & 71.85 & 74.65 & 76.88 & 78.42\\ 
        
        \bottomrule
    \end{tabular}
    }
    \label{tab:nih_auc}
\end{table}

\begin{table}[h!]
\setlength{\tabcolsep}{4.4pt}
    \centering
    \caption{Performance comparison of different methods at various label percentages on the ISIC 2019 dataset, our 20\% test set split. We only show our reproduction without ImageNet pretrainig. Extended results including pretraining or values reported by other works are provided in \cref{app:results}. }
    \small
    \begin{tabular}{llcccc}
        \toprule
        \multirow{2}{*}{Method Type} & \multirow{2}{*}{Method} & \multicolumn{4}{c}{Label Percentage} \\
        \cmidrule(r){3-6}
        & & 2\% & 5\% & 10\%  & 20\% \\
        \midrule
        \multirow{1}{*}{Baseline} 
        & Densenet \cite{DenseNet2017} & 69.37 & 75.35 & 80.39 & 83.39  \\
        \midrule
        \multirow{1}{*}{Consistency} 
        & S2MTS2 \cite{s2mts2_liu2021selfsupervised}  &74.59 & 76.81  & 81.72  &  84.06  \\
        \midrule
        \multirow{1}{*}{Pseudo Label} 
        & Fixmatch \cite{fixmatch_sohn2020fixmatch} & 70.83 & 78.06 & 80.89& 83.76\\

         & ACPL \cite{acpl_liu2022acpl} & 72.35 & 78.47  & 83.69 & 86.57 \\

        \midrule
        \multirow{1}{*}{Diffusion} & \textbf{\ours{}} & 79.11 & 82.03 & 85.31 & 88.83  \\
        \bottomrule
    \end{tabular}
    \label{tab:isic_auc}
\end{table}

\begin{table}[h!]
    \centering
    \setlength{\tabcolsep}{4.4pt}
    \caption{Classification gains from diffusion. Performance comparison of different methods at various label percentages on the ChestXRay14 dataset. We only show results without ImageNet pretrainig. Extended results including pretraining are provided in \cref{app:results}.}
    \small
    \scalebox{1}{
    \begin{tabular}{lccccc}
        \toprule
        \multirow{2}{*}{Method Type} & \multicolumn{5}{c}{Label Percentage} \\
        \cmidrule(r){2-6}
        &  2\% & 5\% & 10\%  & 20\% & 100\%\\
        \midrule
        \multirow{1}{*}{DenseNet \cite{DenseNet2017}} 
        & 64.42 & 66.33  & 69.48 & 74.07 & 74.86   \\

        \midrule
        \multirow{1}{*}{UNet w/o diffusion} 
        & 64.94 & 69.97 & 71.68  & 73.75 & 77.49  \\
        \midrule
        \textbf{\ours{}} & 71.85 &  74.65& 76.88 & 78.43  &  80.59\\
        \bottomrule
    \end{tabular}
    }
    \vspace{-0.1cm}
    \label{tab:nih_supervised}
\end{table}

\vspace{0.1cm}

\Cref{tab:isic_auc} shows the results for the classification task on the ISIC2019 dataset. \ours{} achieves performance comparable to the state-of-the-art ACPL. The gains from \ours{} are again most visible with a lower percentage of labeled data, similar to the results on ChestXRay14.

To further highlight the benefits of our approach, in \Cref{tab:nih_supervised}, we additionally compare the performance of our method to the simple baselines of different architectures. We can see that thanks to the additional diffusion loss, our method can learn more general data representations and achieve higher performance.

\subsection{Visual Counterfactual Explanations}
\label{sec:vce}
\begin{figure}[t]
    \centering
    \includegraphics[width=0.95\columnwidth]{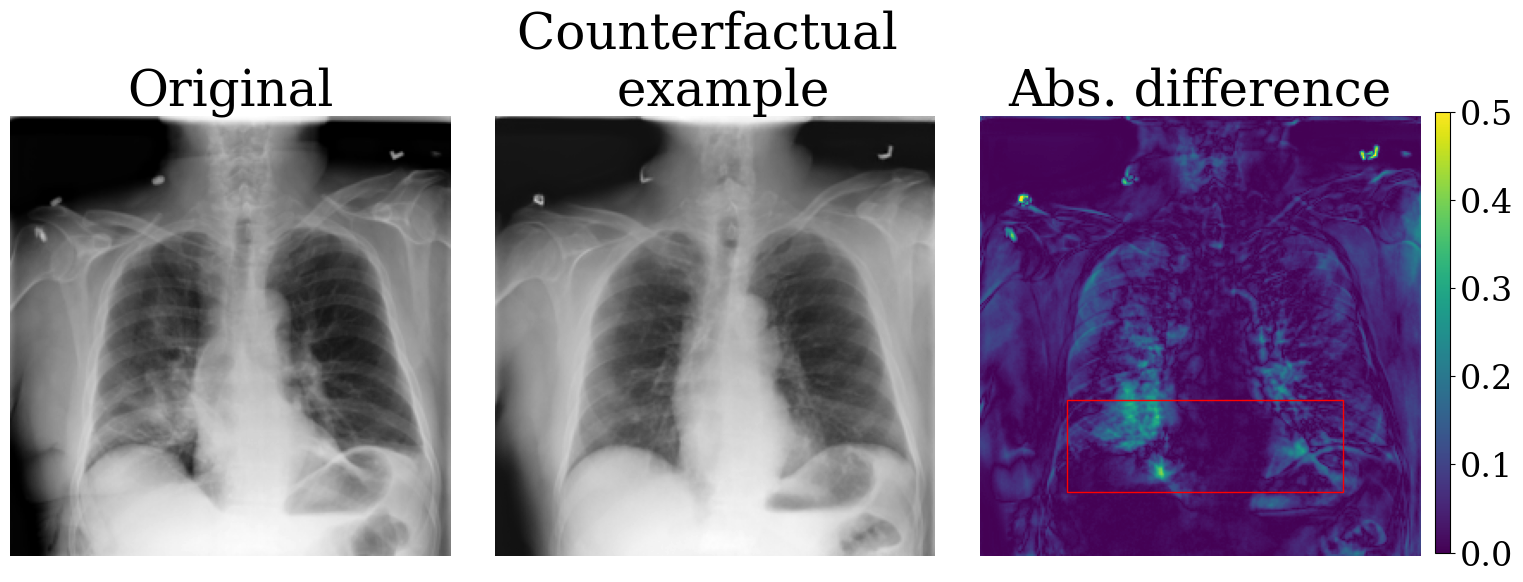}
    \includegraphics[width=0.95\columnwidth]{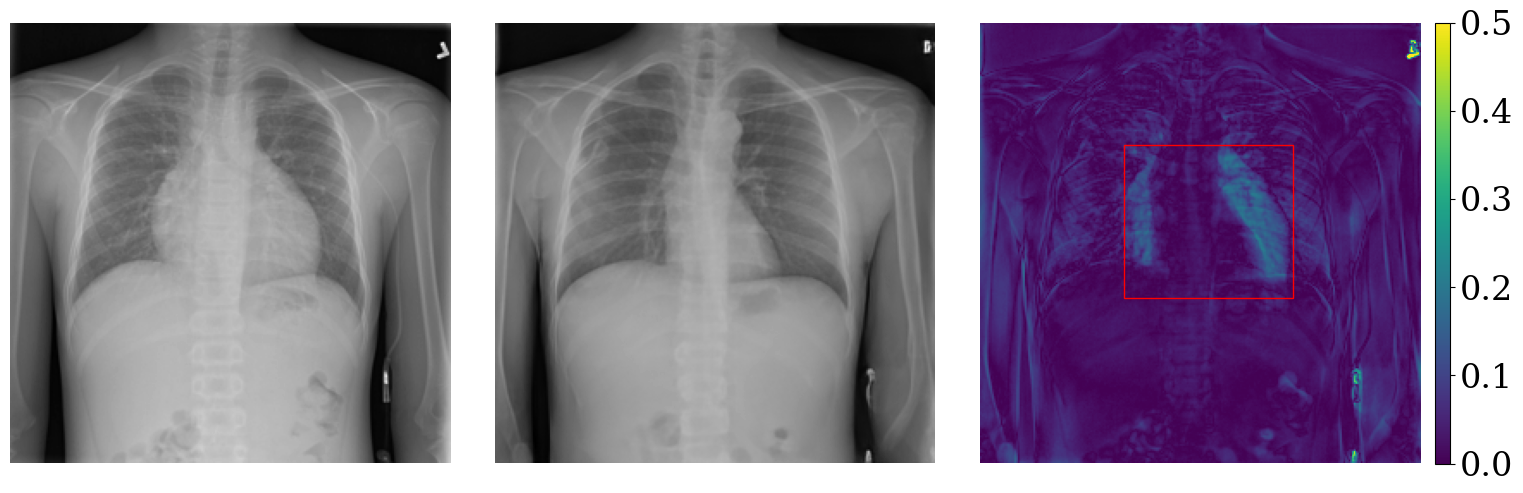}
    \includegraphics[width=0.95\columnwidth]{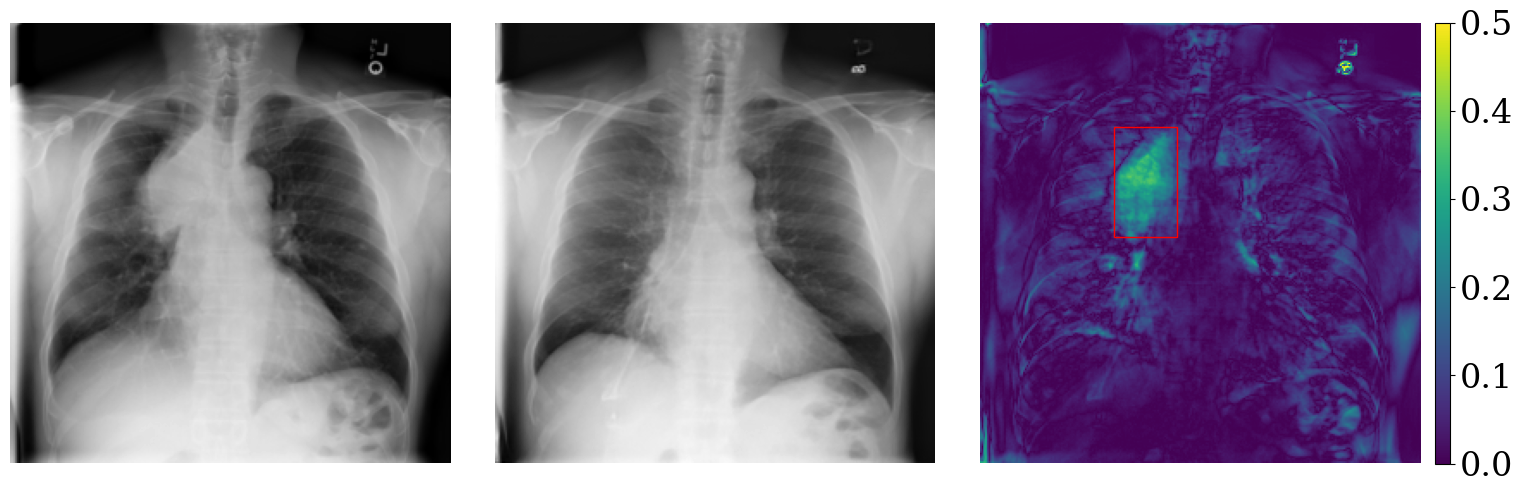}

     \caption{\textbf{Counterfactual examples}. We modify the images to reduce the probability of disease prediction across various classes. From top to bottom: Atelectasis, Cardiomegaly, Mass. The bounding boxes annotated by experts highlight regions that indicate respective diseases.}
    \label{fig:cardiomegaly_remove}
    \vspace{-0.7cm}
\end{figure}

\begin{table}[h]
    \centering
    \small
    \setlength{\tabcolsep}{4pt}
    \scalebox{1}{
    \begin{tabular}{lcccc}
        \toprule
        & \textbf{Orig} & \textbf{Counterfactual} & \textbf{Diff$\downarrow$} & \textbf{Other diff$\downarrow$} \\
        \midrule
        \textbf{Atelectasis} & 0.70 & 0.30 & -0.40 & 0.10 \\
        \textbf{Cardiomegaly} & 0.82 & 0.26 & -0.56 & 0.06 \\
        \textbf{Effusion} & 0.77 & 0.16 & -0.61 & 0.13 \\
        \textbf{Infiltration} & 0.66 & 0.40 & -0.26 & 0.14 \\
        \textbf{Mass} & 0.75 & 0.32 & -0.43 & 0.10 \\
        \textbf{Nodule} & 0.70 & 0.33 & -0.37 & 0.06 \\
        \textbf{Pneumonia} & 0.62 & 0.50 & -0.12 & 0.05 \\
        \textbf{Pneumothorax} & 0.81 & 0.60 & -0.21 & 0.05 \\
        \bottomrule
    \end{tabular}
    }
    \caption{\textbf{Counterfactual examples, disease indication removal. (2\% labeled data)} Average score of the externally trained  CheXNet \cite{chexnet} classifier, when predicting original data samples and counterfactual examples generated with \ours{}. The mean prediction confidence decreases (Diff), while we observe a significantly lower mean absolute prediction change for the remaining classes (Other diff). Notably, this effect is consistent for models trained with both 2\% and 20\% (in \cref{app:results}) of labeled data, indicating that our method extracts meaningful representations 
    even with extremely limited labelling.} 
    \label{tab:classifier_guidance_removing_2p}
\end{table}

\begin{figure}[h]
    \centering
    \includegraphics[width=0.99\columnwidth,trim=0cm 0cm 0cm 0cm,clip]{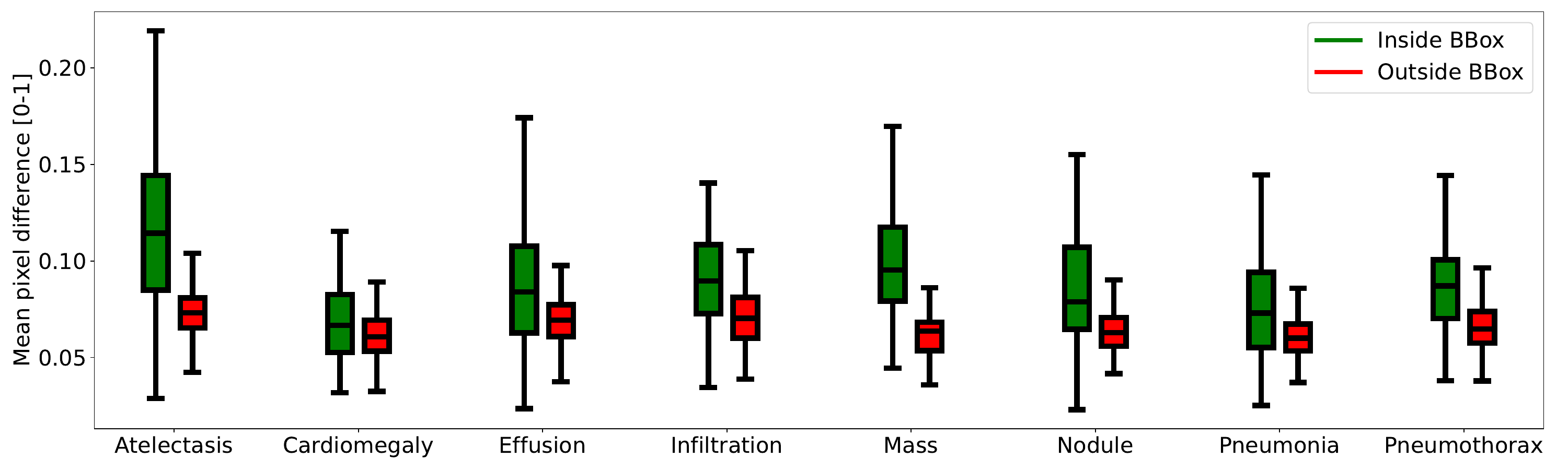}
   \caption{\textbf{Mean pixel differences (0-1) inside and outside the ground-truth bounding boxes when generating counterfactual examples.} For all diseases, the majority of changes of our counterfactual examples generation method occur within the ground-truth boxes (green) assigned by the trained physicians.}
    \label{fig:boxplots_2p}
\end{figure}

\begin{table}[h]
    \centering
    \small
    \setlength{\tabcolsep}{4pt}
    \scalebox{1}{
    \begin{tabular}{lcccc}
        \toprule
        & \textbf{Orig} & \textbf{Counterfactual} & \textbf{Diff $\uparrow$} & \textbf{Other diff$\downarrow$} \\
        \midrule
        \textbf{Atelectasis} & 0.48 & 0.73 & 0.25 & 0.07 \\
        \textbf{Cardiomegaly} & 0.22 & 0.68 & 0.46 & 0.07 \\
        \textbf{Effusion} & 0.45 & 0.84 & 0.39 & 0.08 \\
        \textbf{Infiltration} & 0.54 & 0.66 & 0.12 & 0.07 \\
        \textbf{Mass} & 0.37 & 0.78 & 0.41 & 0.12 \\
        \textbf{Nodule} & 0.40 & 0.74 & 0.34 & 0.13 \\
        \textbf{Pneumonia} & 0.44 & 0.67 & 0.23 & 0.13 \\
        \textbf{Pneumothorax} & 0.34 & 0.86 & 0.52 & 0.15 \\
        \bottomrule
    \end{tabular}
    }
       \caption{\textbf{Classifier guidance, disease indications enforcing. (2\% labeled data)} 
    Average CheXNet~\cite{chexnet} confidence score for original healthy data examples and generated counterfactual examples, when guided towards particular diseases. Our method is able to accurately input disease indicators (diff) without significantly affecting probability of predicting other classes (Other diff).
    } \label{tab:classifier_guidance_enforcing_2p}
\end{table}

An important challenge in medical image analysis is the need for explainability of the methods. In this work, we propose to address it through visual counterfactual explanations that can be generated with our model. The goal of this approach is to explain the classifier's decision by showing which modifications to the original images are needed to alter the prediction. \ours{} allows for creating counterfactual examples with the generative part of our model by noising the original image and denoising it back with guidance towards the opposite target. The successful counterfactual generation method should aim for two objectives:
(1) After modification, each example should have a significantly lower probability of being assigned to its original class.
(2) The modification should occur only in the area related to the evaluated target, e.g., specific to the particular disease.

We run two experiments with the  ChestXRay14 dataset to evaluate the performance of our \ours{} with respect to those objectives. In order to check how well the generative part of our model can remove traces of diseases in the X-ray photos, we use the external classifier to predict the score for original test examples before and after manipulating the image. As presented in Table~\ref{tab:classifier_guidance_removing_2p} and \ref{tab:classifier_guidance_removing_20p} (\cref{app:results}) , when guiding away from the original label, we can observe a significant drop in the probability of the target class. At the same time, we do not observe substantial changes in the predictions for other classes. We visualize several examples of counterfactual examples generation generated with \ours{} in Figure~\ref{fig:cardiomegaly_remove}, where we alter the image to reduce the probability of predicting different diseases. We plot the original images, their counterfactual generations, and differences to highlight applied changes.

\begin{figure}[t!]
    \centering
    \includegraphics[width=0.95\columnwidth]{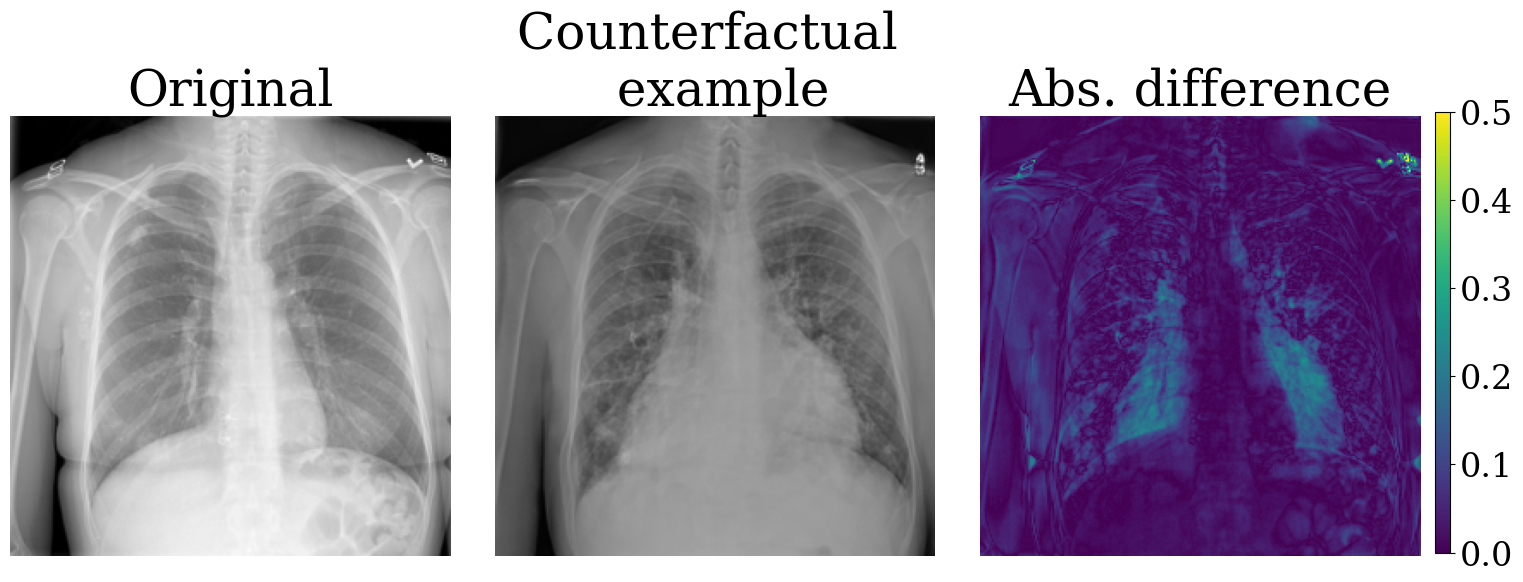}
    \includegraphics[width=0.95\columnwidth]{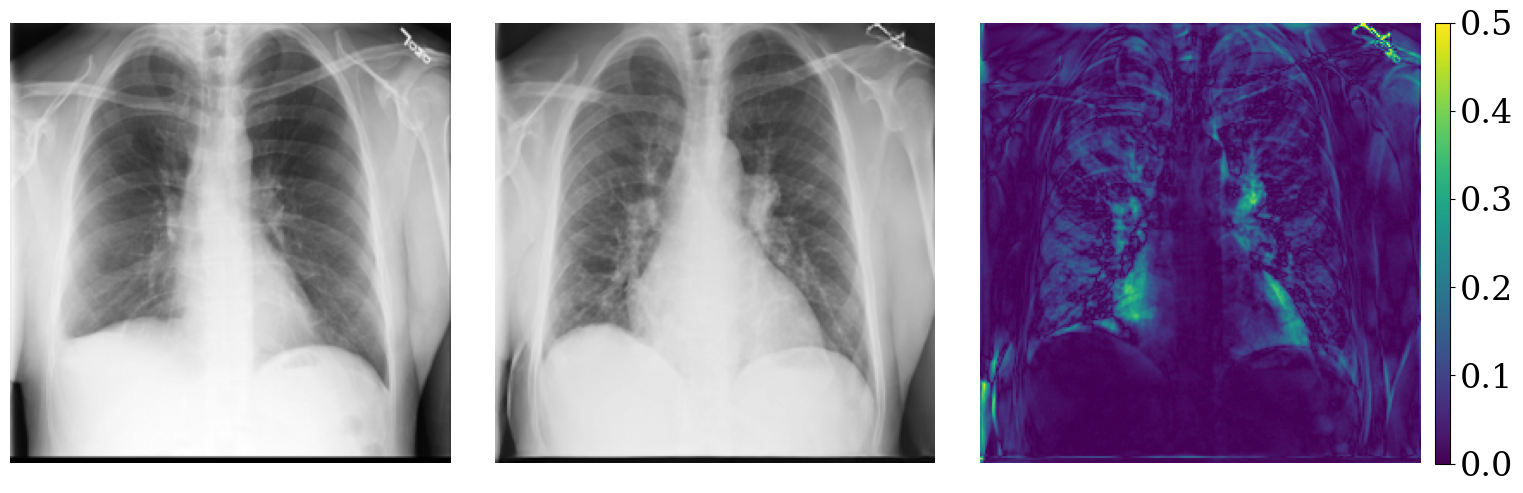}
    \caption{\textbf{Examples of enforcing disease indicators.} Visualisation of counterfactual images generated for healthy data samples that are modified towards cardiomegaly disease characterized by an enlarged heart.} 
    \label{fig:cardiomegaly_enforcing}
    \vspace{-0.7cm}
\end{figure}

To measure the accuracy of the changes performed with our VCE technique, we relate to the ground-truth bounding boxes provided with the selected classes of the  ChestXRay14 dataset surrounding the area of changes indicating the labeled disease. As an evaluation, we check how much the value of pixels located within bounding boxes changed when compared with those outside of the box. We present the mean absolute differences for our two semi-supervised models in Figures Fig.~\ref{fig:boxplots_2p} and Fig.~\ref{fig:boxplots_20p} (\cref{app:results}). As visible for the majority of analyzed diseases, the changes are located within the bounding box. The difference is less visible only in the Cardiomegaly case, where bounding boxes encompass the whole heart, including its non-affected part.

We further extend those experiments to the opposite direction. In Figure~\ref{fig:cardiomegaly_enforcing}, we show how we can use our counterfactual generations method to input the disease indicators to the healthy X-ray scan. We present the quantitative evaluation of this approach in Table~\ref{tab:classifier_guidance_enforcing_2p} and ~\ref{tab:classifier_guidance_enforcing_20p} (\cref{app:results}), where we can observe higher prediction scores for the examples guided towards target diseases. Once more, we show that our VCE method is accurate even in a low-label percentage regime.

\begin{figure}[h!]
    \centering
    \scalebox{0.98}{
    \includegraphics[width=1\columnwidth]{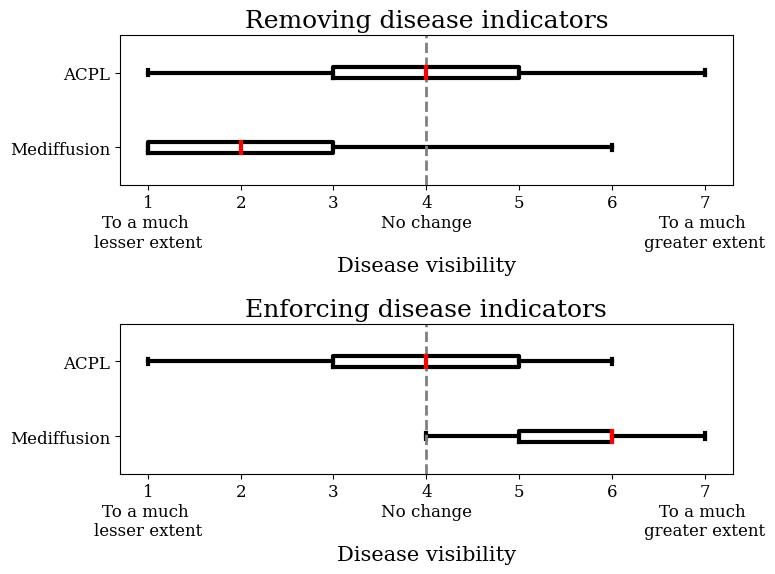}
    }
    \caption{\textbf{Domain experts evaluation of counterfactual examples} The figure shows the results of a survey in which medical doctors rated the visibility of disease indicators in counterfactual images generated by \ours{} and with guidance from external classification trained with ACPL. We compare models trained with labels available for 2\% of samples. The top box plot represents the assessment for removing disease indicators, and the bottom plot represents the assessment for enforcing disease indicators. Doctors rated the counterfactual examples created with \ours{} as demonstrating more noticeable changes in disease indicators in both scenarios.}
    \label{fig:poll}
    \vspace{-0cm}
\end{figure}

Finally, to assess the practical usability of \ours{} in generating VCE we survey 7 medical doctors. We compared our method to counterfactual examples created with guidance from an external classifier trained using ACPL \cite{acpl_liu2022acpl} (see \cref{app:poll} for details). We include 10 examples of enforcing disease indicators and 10 examples of removing disease indicators for each method,  resulting in a total of 280 assessments of original and counterfactual image pairs by domain experts. Participants were asked to rate the extent to which the condition from the original image is visible on the counterfactual example. The counterfactual examples created using \ours{} were rated by doctors as showing more noticeable changes in disease indicators, both when removing and enforcing these indicators, as seen in Figure \ref{fig:poll}. Additionally, we interviewed the participants to gather their preferences regarding the use of artificial intelligence as a decision support tool in clinical practice. The majority of experts, with one exception, expressed a preference for models that offer greater explainability, even at the expense of reduced accuracy, over more accurate models that provide limited explainability.

\subsection{Generation with Classifier Guidance}
\begin{figure}[h!]
    \centering
    \scalebox{0.9}{
    \includegraphics[width=\columnwidth]{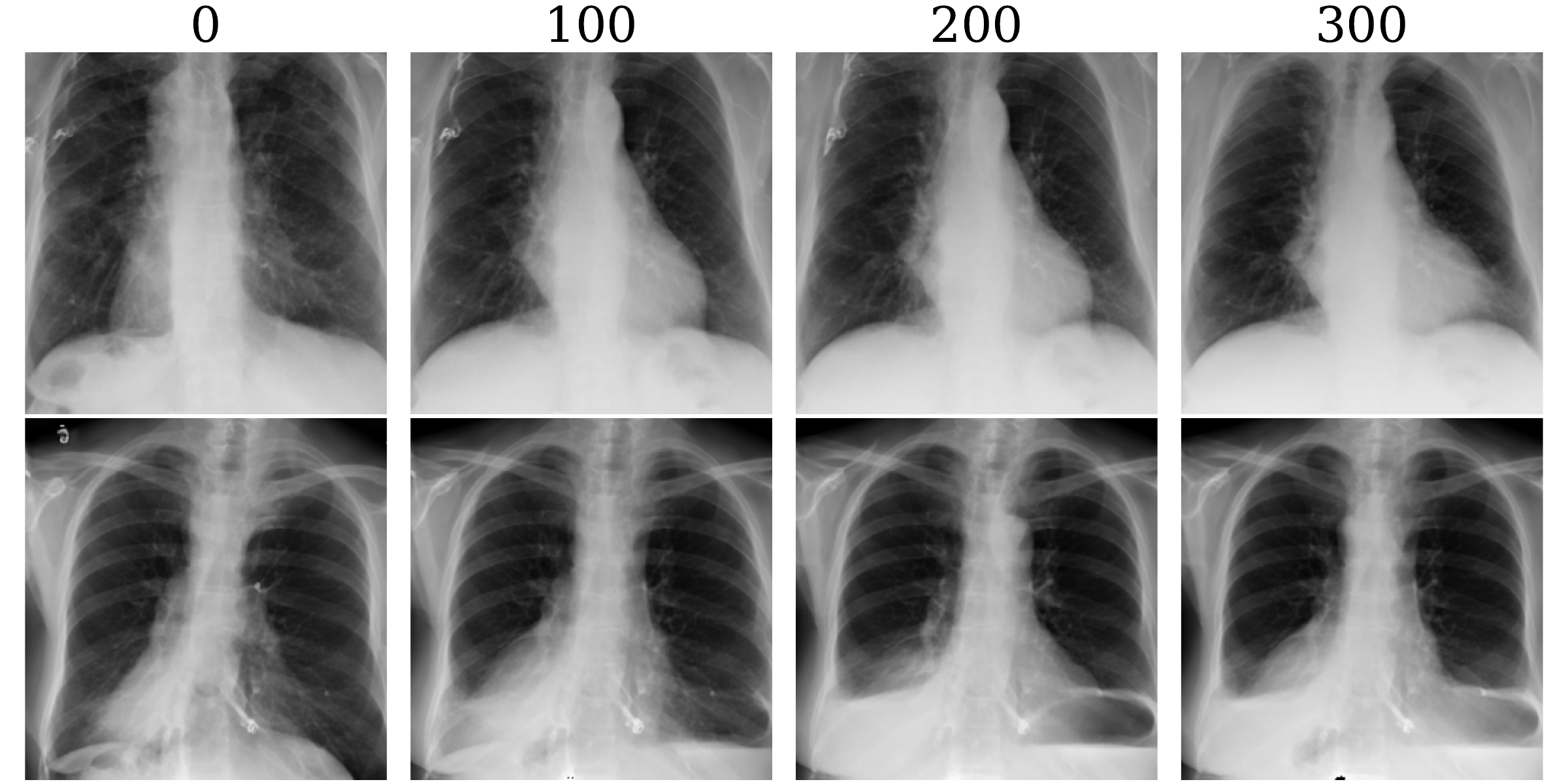}
    }
    \caption{\textbf{Examples of generations with increasing classifier guidance strength.} Top: cardiomegaly disease, bottom: effusion. We can observe that higher classifier guidance strength lead to more prominent disease features without significantly altering other parts of the image.}
    \label{fig:progress}
\vspace{-0.1cm}
\end{figure}

\ours{} facilitates synthetic data generation, naturally leveraging the classifier component to guide the generations towards the desired class label distribution. In \Cref{fig:progress}, we present qualitative examples of samples generated under the guidance of the integrated classifier towards a particular disease. Increasing the strength of the classifier guidance results in more prominent disease features visible on the generated samples, what can be understood as increased severity of the disease. Evaluation of generative capabilities are provided in \cref{app:results}.

\section{Conclusions}

We introduce \ours{}, a method for self-explainable semi-supervised learning with a joint diffusion model in medical domain. Our approach effectively utilizes a shared parameterization to learn from both labeled and unlabeled data, enhancing classifier decisions explainable through the generation of counterfactual examples. In our validation, \ours{} matches current semi-supervised methods' performance while adding robust explainability and the capability to generate synthetic data.

\clearpage
{\small
\bibliographystyle{ieee_fullname}
\bibliography{egbib}
}

\clearpage
\appendix

\section*{Supplementary Material}
This is a supplementary material to our submission Mediffusion: Joint Diffusion for Medical Image Generation and Self-Explainable Semi-Supervised Classification. In this document we describe configuration details for the conducted experiments. In addition, to facilitate reproduction, we attach the code repository.

\section{Additional Results}
\label{app:results}
\begin{table}[h]
    \centering
    \small
    \setlength{\tabcolsep}{4pt}
    \scalebox{1}{
    \begin{tabular}{lcccc}
        \toprule
        & \textbf{Orig} & \textbf{Counterfactual} & \textbf{Diff$\downarrow$} & \textbf{Other diff$\downarrow$} \\
        \midrule
        \textbf{Atelectasis} & 0.70 & 0.28 & -0.42 & 0.12 \\
        \textbf{Cardiomegaly} & 0.82 & 0.30 & -0.52 & 0.08 \\
        \textbf{Effusion} & 0.77 & 0.12 & -0.66 & 0.14 \\
        \textbf{Infiltration} & 0.66 & 0.33 & -0.33 & 0.14 \\
        \textbf{Mass} & 0.75 & 0.29 & -0.47 & 0.09 \\
        \textbf{Nodule} & 0.70 & 0.37 & -0.34 & 0.06 \\
        \textbf{Pneumonia} & 0.62 & 0.54 & -0.08 & 0.03 \\
        \textbf{Pneumothorax} & 0.81 & 0.53 & -0.29 & 0.08 \\
        \bottomrule
    \end{tabular}
    }
    \caption{\textbf{Counterfactual examples, disease indication removal (20\% labeled data).} Average CheXNet classifier score for original and counterfactual samples. The mean prediction confidence decreases (Diff), while changes in remaining classes remain minimal (Other diff).}
    \label{tab:classifier_guidance_removing_20p}
\end{table}

\begin{figure*}[h]
    \centering
    \includegraphics[width=0.8\textwidth,trim=0cm 0cm 0cm 0cm,clip]{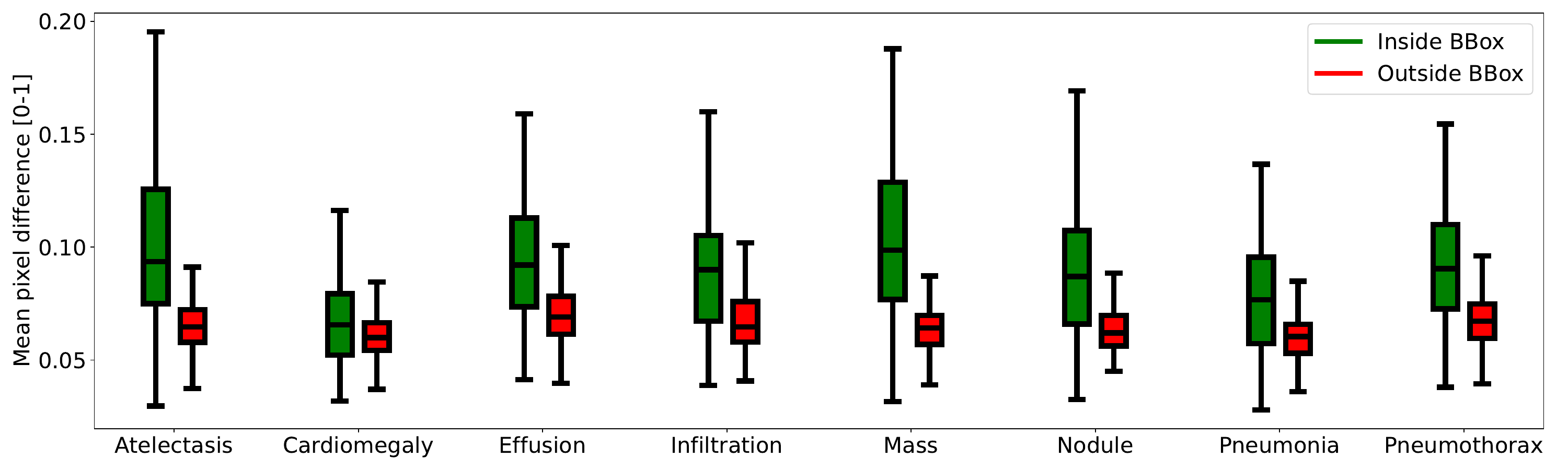}
    \caption{\textbf{Mean pixel differences (0-1) inside and outside the ground-truth bounding boxes (20\% labeled data).} The majority of changes in counterfactual examples occur within the bounding boxes assigned by trained physicians.}
    \label{fig:boxplots_20p}
\end{figure*}

\begin{table}[h]
    \centering
    \small
    \setlength{\tabcolsep}{4pt}
    \scalebox{1}{
    \begin{tabular}{lcccc}
        \toprule
        & \textbf{Orig} & \textbf{Counterfactual} & \textbf{Diff $\uparrow$} & \textbf{Other diff$\downarrow$} \\
        \midrule
        \textbf{Atelectasis} & 0.48 & 0.79 & 0.31 & 0.09 \\
        \textbf{Cardiomegaly} & 0.22 & 0.74 & 0.52 & 0.10 \\
        \textbf{Effusion} & 0.45 & 0.89 & 0.44 & 0.07 \\
        \textbf{Infiltration} & 0.54 & 0.75 & 0.21 & 0.10 \\
        \textbf{Mass} & 0.37 & 0.90 & 0.53 & 0.12 \\
        \textbf{Nodule} & 0.40 & 0.85 & 0.45 & 0.14 \\
        \textbf{Pneumonia} & 0.44 & 0.70 & 0.26 & 0.12 \\
        \textbf{Pneumothorax} & 0.34 & 0.88 & 0.54 & 0.15 \\
        \bottomrule
    \end{tabular}
    }
    \caption{\textbf{Classifier guidance, disease indications enforcing (20\% labeled data).} Average CheXNet confidence scores for original healthy data examples and counterfactuals guided towards disease indications. Minimal impact on other classes (Other diff).}
    \label{tab:classifier_guidance_enforcing_20p}
\end{table}

In this section we present extended results for classification evaluation (\cref{tab:nih_auc_app}, \cref{tab:isic_auc_app}, \cref{tab:nih_supervised_app}).

In \cref{tab:fid_table}, we demonstrate the effectiveness of incorporating classifier guidance in generating data samples that follow a target distribution. To that end, we measure the Fréchet Inception Distance (FID) \cite{fid} and Kernel Inception Distance (KID) \cite{kid} for generated samples. The results demonstrate that guidance from the jointly trained classifier leads to a data distribution that more closely matches the desired target one.

\begin{table}[h]
    \small
    \setlength{\tabcolsep}{4.4pt}
    \centering
    \caption{Performance comparison of different methods at various label percentages on ChestXRay14 dataset at official test split. * denotes our reproduction, \dag~ denotes without ImageNet pretrainig and \ddag~denotes using ImageNet pretraining.}
    \small
    \scalebox{1}{
    \begin{tabular}{llcccc}
        \toprule
        \multirow{2}{*}{Method Type} & \multirow{2}{*}{Method} & \multicolumn{4}{c}{Label Percentage} \\
        \cmidrule(r){3-6}
        & & 2\% & 5\% & 10\% &  20\% \\
        \midrule
        \multirow{2}{*}{Baseline} 
        & Densenet \dag & 64.42 & 66.33  & 69.48 &  74.07   \\
        & Densenet \ddag & 70.27 & 76.14 & 78.08 &79.78    \\
        
        \midrule
        \multirow{3}{*}{Consistency} 
        & SRC-MT~\cite{src_mt_liu2020semi} & 66.95 & 72.29 & 75.28  & 79.76 \\
        & NoTeacher~\cite{noteacher_Unnikrishnan_2021} & 72.60 & 77.04 & 77.61 &  79.49 \\
        &S2MTS2*& 62.22 &65.97&71.01&74.78\\
        \midrule
        \multirow{3}{*}{Pseudo Label}
        & Graph XNet~\cite{graph_xnet_avilesrivero2020graphxnet} & 53.00 & 58.00 & 63.00 & 78.00 \\
        & ACPL* \dag ~\cite{acpl_liu2022acpl} & 72.51 & 77.10 & 79.08 &  80.47 \\
        & ACPL* \ddag ~\cite{acpl_liu2022acpl} & 63.22 & 66.68 & 71.79 &  74.93 \\
        & PEFAT~\cite{pefat_zeng2023pefat} & 75.06 & 79.54 & 80.93  & 82.58 \\
        \midrule
        \multirow{1}{*}{Diffusion} & \textbf{\ours{}} & 71.85 & 74.65 & 76.88 & 78.42\\ 
        
        \bottomrule
    \end{tabular}
    }
    \label{tab:nih_auc_app}
\end{table}

\begin{table}[h!]
\setlength{\tabcolsep}{4.4pt}
    \centering
    \caption{Performance comparison of different methods at various label percentages on the ISIC 2019 dataset, our 20\% test set split. \dag~ denotes no ImageNet pretrainig, \ddag~denotes using ImageNet pretraining. * denotes our reproduction. }
    \small
    \begin{tabular}{llcccc}
        \toprule
        \multirow{2}{*}{Method Type} & \multirow{2}{*}{Method} & \multicolumn{4}{c}{Label Percentage} \\
        \cmidrule(r){3-6}
        & & 2\% & 5\% & 10\%  & 20\% \\
        \midrule
        \multirow{2}{*}{Baseline} 
        & Densenet\dag & 69.37 & 75.35 & 80.39 & 83.39  \\
        & Densenet\ddag  & 79.17 & 84.25 & 87.64 & 91.04 \\
        \midrule
        \multirow{1}{*}{Consistency} 
        & S2MTS2*  &74.59 & 76.81  & 81.72  &  84.06  \\
        \midrule
        \multirow{4}{*}{Pseudo Label} 
        & Fixmatch*\dag & 70.83 & 78.06 & 80.89& 83.76\\
        & Fixmatch*\ddag & 68.89 & 77.83 & 79.63 & 81.47 \\

         & ACPL*\dag \cite{acpl_liu2022acpl} & 72.35 & 78.47  & 83.69 & 86.57 \\
         & ACPL*\ddag \cite{acpl_liu2022acpl} & 79.58 & 85.84  & 88.18   & 92.69 \\

        \midrule
        \multirow{1}{*}{Diffusion} & \textbf{\ours{}} & 79.11 & 82.03 & 85.31 & 88.83  \\
        \bottomrule
    \end{tabular}
    \label{tab:isic_auc_app}
\end{table}

\begin{table}[h!]
    \centering
    \setlength{\tabcolsep}{4.4pt}
    \caption{Classification gains from diffusion. Performance comparison of different methods at various label percentages on the ChestXRay14 dataset. \dag~ denotes no ImageNet pretrainig, \ddag~denotes using ImageNet pretraining.}
    \small
    \scalebox{1}{
    \begin{tabular}{lccccc}
        \toprule
        \multirow{2}{*}{Method Type} & \multicolumn{5}{c}{Label Percentage} \\
        \cmidrule(r){2-6}
        &  2\% & 5\% & 10\%  & 20\% & 100\%\\
        \midrule
        \multirow{1}{*}{DenseNet \dag} 
        & 64.42 & 66.33  & 69.48 & 74.07 & 74.86   \\
        \multirow{1}{*}{DenseNet \ddag} 

                &  70.27 & 76.14 & 78.08 & 79.78 &  81.51 \\
        \midrule
        \multirow{1}{*}{UNet w/o diffusion} 
        & 64.94 & 69.97 & 71.68  & 73.75 & 77.49  \\
        \midrule
        \textbf{\ours{}} & 71.85 &  74.65& 76.88 & 78.43  &  80.59\\
        \bottomrule
    \end{tabular}
    }
    \vspace{-0.1cm}
    \label{tab:nih_supervised_app}
\end{table}

\begin{table}[h!]
\centering
\small
\scalebox{1}{
\begin{tabular}{ccc}
\toprule
 & FID $\downarrow$ & KID $\downarrow$ \\ 
 \midrule
Unguided & 84.81 & 0.0803 \\ 
Guided &  79.22 & 0.0610\\ 
\bottomrule
\end{tabular}
}
\caption{\textbf{Comparing FID and KID values for unconditional and guided generation from the ChestXRay14 dataset.} Results are presented for a model trained with 20\% labeling. We calculate FID and KID on 300 samples from the test split for each class.}
\vspace{-0.4cm}
\label{tab:fid_table}
\end{table}

\section{Extended related work}
\label{app:full_related_work}

\subsection{Diffusion models}

\textbf{Diffusion models in medical domain.}
Diffusion models in the medical domain have been primarily leveraged for synthetic data generation \cite{chen2024generalizable, zingarini2023m3dsynth, ye2023synthetic, txurio, Dorjsembe2023ConditionalDM, kimboah, skorupko2024} and refinement through techniques such as image-to-image translation 
\cite{Kim_2024_WACV, ozbey, kaleta2024}, denoising 
\cite{xiangddm}, super-resolution \cite{wang2023inversesr3dbrainmri}, 
relighting \cite{chen2024lightdiffsurgicalendoscopicimage} and data reconstruction \cite{pengcheng, song2022solving, jalal2021robust}. 
Beyond their traditional applications in generative tasks, diffusion models have also been employed as backbone components for tasks including segmentation \cite{wang2023towards, rahman, zbinden, kim2022diffusion}, anomaly detection \cite{wolleb2022diffusion}, and classification \cite{yang2023diffmic}. 
However, the potential of diffusion models to simultaneously address multiple problems in the medical domain has been relatively understudied. In this vein of research, existing works explore combining image super-resolution \cite{wang2024implicit, xu2024simultaneous, xu2024simultaneous} or segmentation \cite{Barbera2022AnatomicallyCC, pinaya} with other applications.
From this perspective, our proposed \ours{} ~is a novel approach that combines the synthetic data generation capabilities of diffusion models with a framework for facilitating semi-supervised learning of a visually explainable classifier. 

\vspace{0.105cm}

\textbf{H-space in diffusion models.} Several methods explore the potential of features extracted by the UNet model trained as a diffusion denoiser. For the generative task, Kwon et al~\cite{kwon2022diffusion} and Park et al.~\cite{park2023understanding}, show that representations in the UNet bottleneck dubbed \emph{H-space} can be used for image manipulation.
In~\cite{baranchuk2021labelefficient}, authors show how we can benefit from features extracted from the DDPMs trained in an unsupervised way for label-efficient image segmentation. This idea is extended to visual correspondence task in~\cite{luo2023dhf}. In this work, we follow the parametrization introduced in~\cite{jointdiffusion}, where H-features are used as an input to the classifier trained in a supervised or semi-supervised way.

\subsection{Semi-supervised learning for classification}

Semi-supervised learning (SSL) involves learning from both unlabeled and labeled data, with the former typically outnumbering the latter.  In this section, we review applications of such methods in the medical domain.

\textbf{Pseudo-labeling.} Pseudo-labeling assigns labels to unlabeled data based on confidence thresholds, starting with labeled samples. Aviles-Rivero et al.'s GraphXNet \cite{graph_xnet_avilesrivero2020graphxnet} employs a graph-based model for X-ray classification to reduce dependency on annotations. FixMatch \cite{fixmatch_sohn2020fixmatch} generates pseudo-labels for weakly augmented images with high-confidence predictions, then trains on their strongly augmented versions. NoisyStudent \cite{noisystudent_Xie_2020_CVPR} iteratively trains student networks on noised inputs with pseudo-labels from updated teacher networks. ACPL \cite{acpl_liu2022acpl} improves pseudo-label accuracy using classifier ensembling and anti-curriculum learning.

\textbf{Consistency regularization.} Consistency regularization encourages prediction stability across different augmented views of unlabeled data. SRC-MT \cite{src_mt_liu2020semi} enforces semantic consistency under perturbations, helping extract information from unlabeled data. S2MTS2 \cite{s2mts2_liu2021selfsupervised} pre-trains a self-supervised teacher using contrastive learning to guide consistent predictions. VAT \cite{vat_miyato} introduces virtual adversarial perturbations to enforce consistency despite image-level adversarial noise, while PEFAT \cite{pefat_zeng2023pefat} extends VAT by applying feature-level adversarial noise and a sample selection strategy for reliable pseudo-labeling.

\textbf{Relation to \ours.}
Our proposed method is orthogonal to 
the semi-supervised classification approaches described in this section. Rather than employing pseudo-labeling strategies or enforcing consistency regularization, our method focuses on jointly learning data representations shared between supervised and unsupervised (generative) tasks. This allows us to utilize the available data effectively.


\subsection{Visual Counterfactual Explanations}
Visual Counterfactual Explanations (VCE) \cite{vce} is a tool for interpreting machine learning models, particularly in image-based tasks such as medical image classification. VCE aims to provide insights into model decisions by generating counterfactual images that depict minimal changes necessary to alter a model’s prediction. This method enhances the transparency of deep learning models by illustrating which input features of an image are analyzed by the classifier. 
Various studies \cite{explenetion1, explenetion2, explenetion3} have demonstrated the effectiveness of VCE in medical imaging by generating realistic visual explanations that are easily interpretable by clinicians. Outside the medical context, the potential of diffusion models in providing visual explanations of model decisions has been demonstrated by \cite{augustin2022diffusion}.
In this work, we extend previous works with joint generative-discriminative modeling which allows for the accurate generation of counterfactual explanations without an external model.

\section{Latent Autoencoder Specification and Configuration}
\label{app:latencoder}

Given an image \( x \in \mathbb{R}^{H \times W \times C} \) in either RGB or grayscale format, the encoder \(\mathcal{E}\) maps \( x \) to a latent representation \( z = \mathcal{E}(x) \). The decoder \(\mathcal{D}\) then reconstructs the image from this latent representation, resulting in \(x' = \mathcal{D}(z) = \mathcal{D}(\mathcal{E}(x)) \), where \( z \in \mathbb{R}^{h \times w \times c} \). In this process, the encoder reduces the resolution of the input sample by a factor of \( f = \frac{H}{h} = \frac{W}{w} \).

We train our autoencoder model using a combination of perceptual loss and a patch-based adversarial objective. This approach ensures that the reconstructions remain within the image manifold, promoting local realism and preventing the blurriness typically caused by pixel-space losses such as $L_2$ or $L_1$. Additionally, we employ the KL-regularization term, which introduces a slight KL divergence penalty towards a standard normal distribution of latent variables, similar to the regularization in VAEs \cite{kingma2014autoencoding}.

By employing our trained perceptual compression model, we create an efficient, low-dimensional latent space that abstracts away high-frequency, imperceptible details. This latent space, in contrast to the high-dimensional pixel space, is better suited for likelihood-based generative models as it enables them to (1) concentrate on significant, semantic features of the data and (2) operate within a lower-dimensional, computationally efficient framework. The autoencoder model is trained in an unconditional manner, allowing us to leverage available unlabeled data. Despite the lack of labels, the encoder model effectively captures features advantageous for both generative and classification tasks.

In our experiments, we follow the implementation of the KL-autoencoder employed by Rombach et al. \cite{rombach2022highresolution}.

\paragraph{Model Architecture}
The base number of channels in the model is set to 128. The channel multiplier factors across different stages are [1, 2, 4, 4], which determine the progression of channel counts in the network. Each stage of the autoencoder includes two residual blocks. The encoder compresses the input sample to a \(32 \times 32 \times 4\) latent representation.

\paragraph{Model Training}
We train the autoencoder model with a base learning rate of \(4.5 \times 10^{-6}\) over 300,000 training steps. The batch size is set to 64. The training images have a resolution of \(256 \times 256\) pixels, with RGB images (\(256 \times 256 \times 3\)) for the ISIC2019 dataset and grayscale images (\(256 \times 256 \times 1\)) for the ChestXRay14 dataset. Perceptual loss is incorporated starting from step 50,000 with a weight of 0.5, and a KL divergence loss is applied with a weight of \(1.0 \times 10^{-6}\).

\section{Diffusion Model Configuration}
For the diffusion model, once more, we build our implementation on the work of Rombach et al. \cite{rombach2022highresolution}, adapting the UNet architecture to incorporate a joint classifier module.

\paragraph{Model Architecture}
In our experiments, we utilize the following UNet architecture. The latent sample size is set to \(32 \times 32 \times 4\). The base number of channels in the convolutional layers is 256. Each stage of the U-Net includes two residual blocks, with the channel count increasing progressively across stages by factors of 1, 2, and 4. Attention mechanisms are incorporated at resolutions of \(4 \times 4\), \(2 \times 2\), and \(1 \times 1\). For the attention mechanism, the model employs a multi-head setup with each head consisting of 32 channels.

We limit the size of the classification features extracted from the UNet model to 10,000 using an average pooling operation. The classifier component is a multilayer perceptron with two fully connected layers. The first layer uses a LeakyReLU activation function with a negative slope of 0.2.

\paragraph{Model Training}
For experiments on the ChestXRay14 dataset with 2\% of labeled samples, we use the Adam optimizer with a base learning rate of \(5.0 \times 10^{-5}\). The learning rate for the classifier is set to \(1.0 \times 10^{-5}\). The classification loss is scaled with a weight of 0.00005, and the classification process begins after 5000 training steps. We train the model for 200,000 steps, using a batch size of 64 for the diffusion component and 32 for the classification component. For scenarios with 5\% of labeled data, we increase the classifier learning rate to \(2.0 \times 10^{-5}\) and the classification loss weight to 0.00015. For higher percentages of labeled data (10\%, 15\%, 20\%, and 100\%), we use a classification loss weight of 0.001 and set the classifier learning rate to \(5.0 \times 10^{-5}\).

For experiments on the ISIC2019 dataset, we use the Adam optimizer with a learning rate of \(5.0 \times 10^{-5}\). The classification loss is scaled with a weight of 0.001. We train the model for 20,000 steps, using a batch size of 64 for the diffusion component and 32 for the classification component. This configuration is used for all percentages of labeled data.

\section{Baseline methods configurations}
\label{app:baseline}
In this section, we provide details regarding our reproduction of baseline methods. Consistently with the experimental setups in the respective original papers, all methods employ the DenseNet121 backbone.

\paragraph{NIH ChestXRay14}
For the experiments with the standard DenseNet121, we utilize the Adam optimizer with a learning rate of 0.03 and a batch size of 16. We apply the same augmentations as in the warm-up phase of ACPL \cite{acpl_liu2022acpl}, resizing the images to 256 $\times$ 256 pixels. The model pretrained on ImageNet is trained for 40 epochs, while the model without pretraining weights is trained for 100 epochs.

For ACPL \cite{acpl_liu2022acpl}, we follow the parameters specified by Liu et al.: Adam optimizer with a learning rate of 0.03, batch size of 16, 20 epochs for the warm-up training, and 10 epochs for each of the 5 ACPL loops. We use the same augmentations as Liu et al., resizing the images to 256 $\times$ 256 pixels. In experiments with ACPL without ImageNet pretraining weights, we use the same parameters but increase the warm-up training to 100 epochs.

For S2MTs2 \cite{s2mts2_liu2021selfsupervised}, we follow the training configurations provided by the authors. The images are resized to 256 $\times$ 256 pixels. The model is pretrained on the entire training dataset in a self-supervised manner for 100 epochs, followed by an additional 10 epochs of fine-tuning as described in the original article.

For experiments involving training the classifier on top of the UNet architecture (without diffusion training) (\cref{tab:nih_supervised}), we use the Adam optimizer with a learning rate of 0.0001, a batch size of 64, and train the model for 100 epochs.

\paragraph{ISIC2019}
For the experiments with the standard DenseNet121, we use the Adam optimizer with a learning rate of 0.001 and a batch size of 32. We apply the same augmentations as described in the warm-up phase of ACPL \cite{acpl_liu2022acpl}, resizing the images to a resolution of 224 $\times$ 224 pixels. For the model pretrained on ImageNet, we train for 40 epochs, while for the model without pretraining weights, we train for 100 epochs.

For the FixMatch experiments \cite{fixmatch_sohn2020fixmatch}, we set the learning rate to 0.0001 and the batch size to 36. The training consists of 100,000 steps and the confidence threshold is set to 0.95.

For ACPL \cite{acpl_liu2022acpl}, we adhere to the parameters specified by Liu et al.: a learning rate of 0.001, a batch size of 32, 40 training epochs for the warm-up training, and 20 training epochs for each of the 5 ACPL training loops. We use the same augmentations as Liu et al., resizing the images to 224 $\times$ 224 pixels. In experiments without ImageNet pretraining weights, we increase the number of warm-up training epochs to 100.

Similarly, for S2MTS2 \cite{s2mts2_liu2021selfsupervised}, we once more follow the training configurations provided by the authors in their code repository. As implemented in original paper, we resize images to a 224 $\times$224 resolution, pretrain the model on the whole training dataset in a self-supervised manner for 100 epochs, with additional 10 epochs of finetuning.

\section{Visual Counterfactual Explanations experiments configurations}
\label{app:vce}
We obtain the counterfactual examples of disease indicators removal visible in \cref{fig:cardiomegaly_remove} and conduct the qualitative evaluation presented in \cref{tab:classifier_guidance_removing_2p}, \cref{fig:boxplots_2p}, \cref{tab:classifier_guidance_enforcing_20p} \cref{fig:boxplots_20p} with the following procedure. First, we add noise to respective class samples running the forward diffusion process up to $t=300$. Then we run the backward diffusion process to $t=0$ with classifier guidance to reduce the score prediction of a given disease class with a guidance scale equal to 500.

For the experiments corresponding to disease indicators enforcing in \cref{tab:classifier_guidance_enforcing_2p}   
and \cref{tab:classifier_guidance_enforcing_20p} we run the forward diffusion process up to $t=200$ and then run the denoising process to maximize the score prediction of a given disease class with classifier guidance scale equal to 200.

\section{Generation with Classifier Guidance experiments configurations}
For the experiments on generating synthetic data in \Cref{tab:fid_table} and \cref{fig:progress} we generate new samples using DDIM sampling method with number of timesteps equal to 50. For the guided generation in \cref{tab:fid_table} we use classifier guidance scale equal to 300. For the generations with progressive gudience scale in \Cref{fig:progress} we use scale values of 0, 100, 200, 300.  

\section{Domain experts evaluation of counterfactual examples}
\label{app:poll}

To evaluate the practical utility of \ours{} for generating Visual Counterfactual Explanations (VCEs), we conduct a survey with seven medical professionals. We discuss the expert evaluation at the end of \cref{sec:vce} and in \cref{fig:poll}. We compare our approach to counterfactual examples generated using an external classifier trained with the ACPL framework \cite{acpl_liu2022acpl}. In both settings, the models are trained on the ChestXRay14 dataset, with access to only 2\% of the available labels.

\textbf{VCEs with an external classifier} We generate counterfactual examples for \ours{} following the procedure outlined in \cref{app:vce}. For the ACPL-based classifier, we apply a corresponding procedure to produce counterfactual examples. The external classifier model is trained with hyperparameters specified in \cref{app:baseline}. To explain the classifier's decision, we use its gradients to guide the diffusion process, employing the same score-based conditioning method used in \ours{}, but replacing the integrated classifier with the ACPL-trained classifier.

Since \ours{} employs a latent diffusion model that operates in the latent space $z_t$, while the ACPL-trained classifier operates in pixel space, we decode $z_t$ back into pixel space $x_t$ using the decoder $D$. This allows us to compute the classifier's gradients for $z_t$.

For counterfactual examples of disease indicator removal, we first add noise to respective class samples running the forward diffusion process up to $t=300$. Then we run the backward diffusion process to $t=0$ with classifier guidance to reduce the score prediction of a given disease class. We use guidance scale equal to 500 for \ours{}. To ensure fairness of comparison, for the external classifier trained with ACPL we use guidance scale 100 larger, which empirically corresponds to the same value of gradient norm.

Similarly, for counterfactual examples of disease indicator enforcing we run the forward diffusion process up to $t=200$ and then run the denoising process to maximize the score prediction of a given disease class. We use classifier guidance scale equal to 200 for \ours and 100 larger for the external classifier trained with ACPL.

\begin{figure}[ht]
    \centering
    \includegraphics[width=\linewidth]{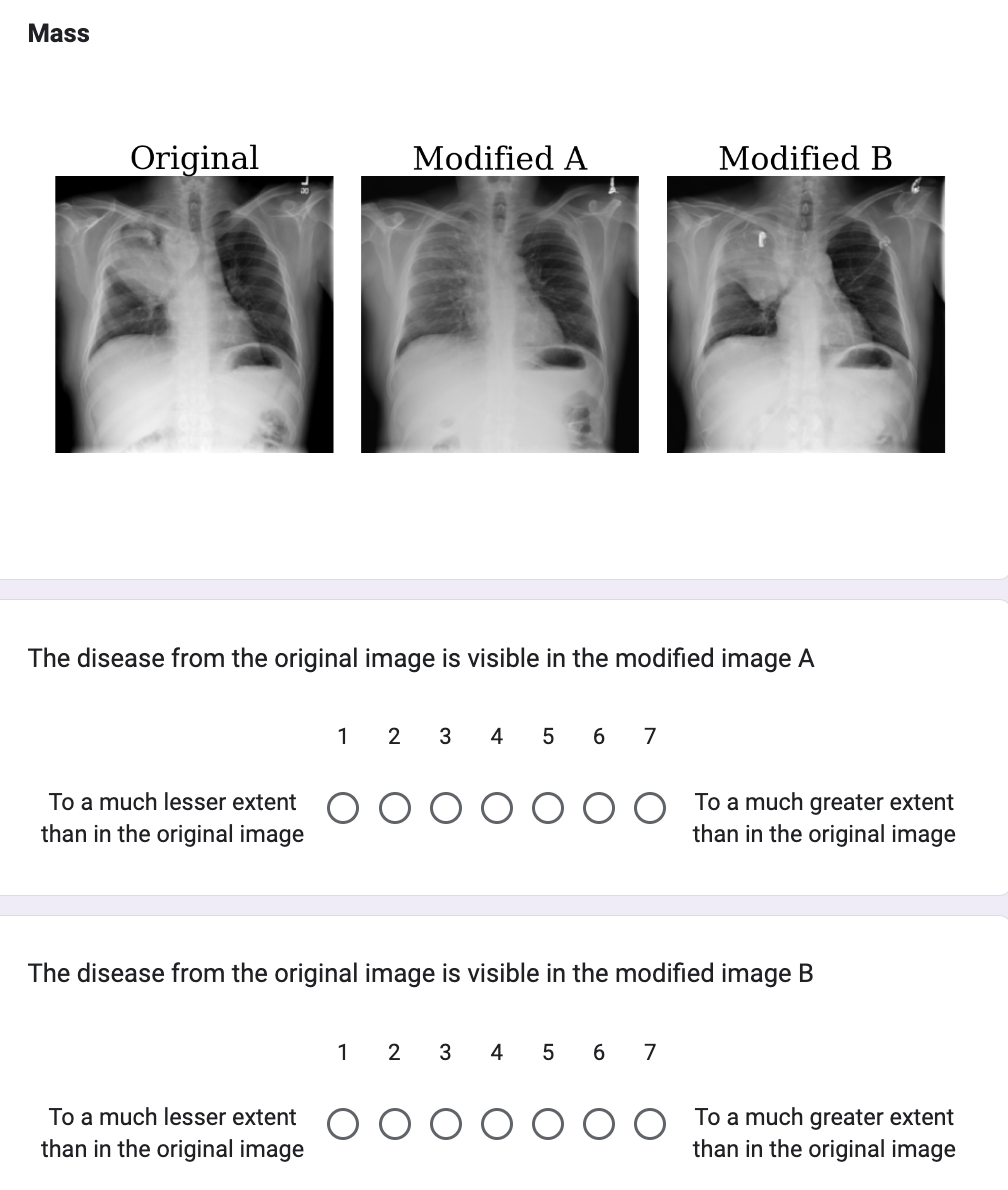}
    \caption{\textbf{Example question for the expert evaluation survey.}}
    \label{fig:example_poll}
\end{figure}

\textbf{Expert evaluation} We include 10 examples of enforcing disease indicators and 10 examples of removing disease indicators for each method,  resulting in a total of 280 assessments of original and counterfactual image pairs by domain experts. Participants were asked to rate the extent to which the condition from the original image is visible on the counterfactual example. We given an example of a survey question on \cref{fig:example_poll}.

\section{Classifier Guidance}
\label{app:classifier_guidance}

Classifier guidance, as introduced by \cite{dhariwal2021diffusion}, is a method for conditioning diffusion models with class labels during generation. In a general case, this approach utilizes a classifier  $p_{\phi}(y|x_t)$ and its gradients  $\nabla_{x_t} \log p_{\phi}(y|x_t)$ to guide the diffusion sampling process towards a particular class label. In line with \cite{dhariwal2021diffusion}, we apply classifier guidance using the score-based conditioning method.

This technique is based on a score-conditioning trick adapted from Song et al. \cite{huang2022improving}, which draws on the relationship between diffusion models and score matching \cite{song2019generative}. In a general case, if we use a model that predicts the noise $\hat{\epsilon}_{\theta}(x_t)$ added to a sample $x_t$, we can derive the following score function:

\begin{equation}
    \nabla_{x_t} \log p_{\theta}(x_t) = - \frac{1}{\sqrt{1 - \bar{\alpha}_t}} \hat{\epsilon}_{\theta}(x_t)
\end{equation}

This expression can be substituted into the score function for $p(x_t)p(y|x_t)$:

\begin{equation}
    \nabla_{x_t} \log(p_{\theta}(x_t)p_{\phi}(y|x_t)) = \nabla_{x_t} \log p_{\theta}(x_t) + \nabla_{x_t} \log p_{\phi}(y|x_t)
\end{equation}

\begin{equation}
    = - \frac{1}{\sqrt{1 - \bar{\alpha}_t}} \hat{\epsilon}_{\theta}(x_t) + \nabla_{x_t} \log p_{\phi}(y|x_t)
\end{equation}

We can then define an updated noise prediction $\hat{\epsilon'}(x_t)$, which corresponds to the score of the joint distribution:

\begin{equation}
    \hat{\epsilon}'(x_t) := \hat{\epsilon}_{\theta}(x_t) -  \sqrt{1 - \bar{\alpha}_t} \nabla_{x_t} \log p_{\phi}(y|x_t)
\end{equation}

This modification enables the use of the standard diffusion sampling procedure, but with adjusted noise predictions \(\hat{\epsilon}'_{\theta}(x_t)\) instead of \(\hat{\epsilon}_{\theta}(x_t)\), guiding the denoised sample towards the desired class $y$.

Specifically, in our joint parametrization framework for \ours{}, we leverage the classifier component \( g_\omega \) of the model to guide the generation process. Furthermore, since our method leverages a latent diffusion model, we replace $x_t$ with its latent representation $z_t$, where \( z = \mathcal{E}(x) \). Therefore, in the case of \ours{} we define the updated noise prediction in the latent space as \(\hat{\epsilon}'(z_t)\): 

\begin{equation}   
    \hat{\epsilon}'(z_t) := \hat{\epsilon}_{\theta}(z_t) - \sqrt{1 - \bar{\alpha}_t} \nabla_{z_t} \log g_\omega(y|e_\nu(z_t))
\end{equation}

\section{Computational resources}
\label{app:compute}

Each experiment reported in this work was run on two Nvidia A100 GPUs.


\end{document}